# Deep learning for image segmentation: veritable or overhyped?

ZHENZHOU WANG*

*College of electrical and electronic engineering, Shandong University of technology, Zibo City, 255000,China*
*wangzz@sdut.edu.cn*

**Abstract:** Deep learning has achieved great success as a powerful classification tool and also made great progress in sematic segmentation. As a result, many researchers also believe that deep learning is the most powerful tool for pixel-level image segmentation. Could deep learning achieve the same pixel-level accuracy as traditional image segmentation techniques by mapping the features of the object into a non-linear function? This paper gives a short survey of the accuracies achieved by deep learning so far in image classification and image segmentation. Compared to the high accuracies achieved by deep learning in classifying limited categories in international vision challenges, the image segmentation accuracies achieved by deep learning in the same challenges are only about 80%. On the contrary, the image segmentation accuracies achieved in international biomedical challenges are close to 95%. Why the difference is so big? Since the accuracies of the competitors' methods are only evaluated based on their submitted results instead of reproducing the results by submitting the source codes or the software, are the achieved accuracies verifiable or overhyped? We are going to find it out by analyzing the working principle of deep learning. Finally, we compared the accuracies of state of the art deep learning methods with a threshold selection method quantitatively. Experimental results showed that the threshold selection method could achieve significantly higher accuracy than deep learning methods in image segmentation.

## 1. Introduction

Most deep learning models are based on artificial neural network that is invented by simulating human brains. Although most deep learning models have various differences from the structural and functional properties of human brains, they are analogical to some extent. The neural dynamics of human brain corresponds to a form of variational EM algorithm, i.e. with approximate rather than exact posteriors [1]. For instance, the human brain could recognize an object without knowing how to calculate its exact size, height, shape or other information and just judge based on some approximated information. Similar to human brains, deep learning models classify the objects without knowing how to calculate its exact size, height, shape or other information. In nature, deep learning combines many continuous regression units discretely and approximates a discrete non-linear mapping function without knowing its exact form. By nature, deep learning is similar to, but different from regression.

The linear regression model constructs the global linear mapping between the inputs and outputs. Different from the linear regression model, deep learning needs to construct the local linear models to determine the global non-linear mapping between the inputs and the outputs. We show a simple example in Fig.1 to demonstrate the difference between the linear regression model and the deep learning model. For simplicity, we only draw one hidden layer of deep learning. For the linear regression model in Fig. 1(a), there are three variables for the inputs and two variables for the outputs and they are formulated by the following equations.

$$y1 = w11 \times x1 + w21 \times x2 + w31 \times x3 + b1 \qquad (1)$$

$$y2 = w12 \times x1 + w22 \times x2 + w32 \times x3 + b2 \qquad (2)$$

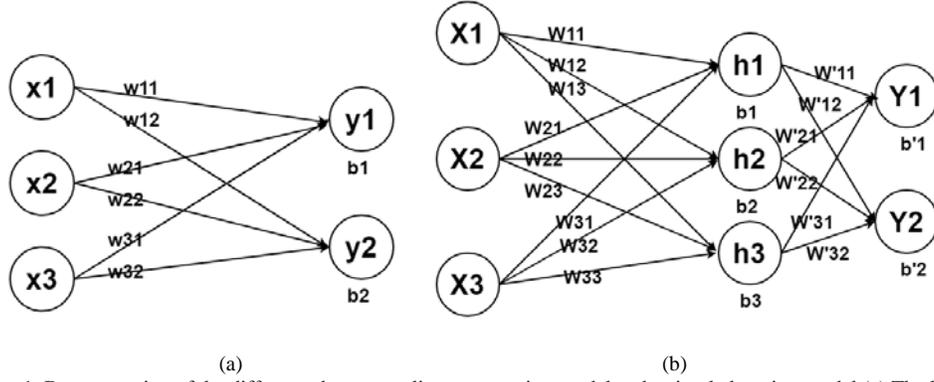

(a) (b)
Fig. 1. Demonstration of the difference between a linear regression model and a simple learning model (a) The linear regression model; (b) The simplified deep learning model.

As can be seen, the above model can be easily solved by the least squares method. However, when the inputs are not linearly separable, there will be no linear mapping between the inputs and the outputs. Deep learning uses many hidden layers to construct the nonlinear mapping. As simplified in Fig. 1 (b), one hidden layer with three neurons is added and the nonlinear mapping between the input and the output is formulated by the following equations.

$$h1 = W11 \times X1 + W21 \times X2 + W31 \times X3 + b1 \tag{3}$$

$$h2 = W12 \times X1 + W22 \times X2 + W32 \times X3 + b2 \tag{4}$$

$$h3 = W13 \times X1 + W23 \times X2 + W33 \times X3 + b3 \tag{5}$$

$$Y1 = W'11 \times h1 + W'21 \times h2 + W'31 \times h3 + b'1 \tag{6}$$

$$Y2 = W'12 \times h1 + W'22 \times h2 + W'32 \times h3 + b'2 \tag{7}$$

A process called 'training' is used to solve the above model and it relies on descent algorithms [2]. Different from the linear regression, deep learning could only determine these parameters to make the outputs respond to the inputs according to the training data instead of generate a continuous and meaningful mapping function between the inputs and the outputs. With the continuous and meaningful mapping function, i.e. the exact mathematical equations, regression could determine the linear or non-linear mapping between the continuous input variables and the continuous output variables. On the contrary, deep learning could only determine the non-linear mapping between the discrete input variables and the finite number of discrete output variables with a set of discrete training parameters.

Without loss of generality, regression is formulated as a mapping model by the following simple equation.

$$y \approx f(x, w) \tag{8}$$

$f$ is a continuous linear or non-linear mapping function and it has the exact mathematical function. $w$ denotes the coefficients to be determined by regression. $x$ denotes the continuous input variable and $y$ denotes the continuous output variable.

Similarly, deep learning could be formulated as:

$$Y_i \approx F(X_j, W); i = 1, \ldots, I; j = 1, \ldots, J \tag{9}$$

$F$ is a discrete and non-linear mapping function and its exact mathematical function is never known. $W$ denotes the discrete mapping parameters to be determined by training. $X_j$ denotes the discrete input variable and $I$ denotes its total number. $Y_i$ denotes the discrete output variable and $J$ denotes its total number.

For image segmentation, the inputs of the deep learning are the features extracted from the image and the output is a prediction map. For the predication map to have pixel-level accuracy, the number of the outputs $M$ is computed as:
$$J = L^{M \times N} \tag{10}$$
where $M$ is the width of the image and $N$ is the height of the image. $L$ denotes the total number of labeling. If the image is segmented into two classes, $L$ equals 2. If the image is segmented into three classes, $L$ equals 3, etc. To segment an image with a resolution of $100 \times 100$ into two classes, the number of outputs is close to infinity, which is intractable for practical implementation. Consequently, the predication map could reach the semantic level in practice. Therefore, traditional image segmentation techniques are usually used to refine the prediction result to achieve the pixel-level accuracy.

## 2. Deep learning for object classification and detection

The deep learning revolution took place around 2012 when it won several well-known international speech and image recognition or classification competitions. The competitions that made deep learning resurge include, but not limited to: large-scale automatic speech recognition competition [3-4], mixed national institute of standards and technology (MNIST) database competition [5], traffic sign competition [6], Chinese handwriting recognition competition [7], ImageNet large scale visual recognition competition [8-9], pattern analysis statistical modeling and computational learning (PASCAL) visual object classes database competition [10] and so on.

Large-scale automatic speech recognition competition uses the Texas Instrument and Massachusetts Institute of Technology (TIMIT) dataset that contains a total of 6300 sentences, 10 sentences spoken by each of 630 speakers from 8 major dialect regions of the United States. The test data has a core portion containing 24 speakers, 2 male and 1 female from each dialect region. According to Eq. (9), the number of inputs is 6300 and the number of the outputs is 8, which is sufficient for deep learning to determine the training parameters robustly. In [4], it is reported that the recognition error rate for the training sets is 13.3 and the recognition error rate for the core test sets is 16.5. The MNIST database of handwritten digits has a training set of 60,000 examples and a test set of 10,000 examples. According to Eq. (9), the number of inputs is 60000 and the number of the outputs is 10, which is sufficient for deep learning to determine the training parameters. In [5], it is reported that the error rate is only 0.23%. Traffic sign competition contains 39209 training images in 43 classes. According to Eq. (9), the number of inputs is 39209 and the number of the outputs is 43, which is sufficient for deep learning to determine the training parameters. In [6], it is reported that a recognition rate of 99.46% had been achieved and it was better than the one of humans on this task (98.84%). Chinese handwriting recognition competition contains a large dataset of isolated characters with about 3.9 million samples of 7,356 classes (7,185 Chinese characters and 171 symbols). According to Eq. (9), the number of inputs is 3900000 and the number of outputs is 7356, which is also sufficient for deep learning to determine the training parameters robustly. In [7], a 97.39% character recognition rate was achieved. ImageNet image classification competition between 2012 and 2014 contains 1281167 images for 1000 classes of objects. According to Eq. (9), the number of inputs is 1281167 and the number of outputs is 1000, which is sufficient for deep learning to determine the training parameters. In [8], an error rate of 15.3% had been achieved, which ignite the revolution of deep learning. In [9], it is reported that the best error rate 6.66% was achieved. However, it only achieved 43.93% average precision for object detection, which already ranked the first among 17 competing teams. In the last session of ImageNet Large Scale Visual Recognition Challenge that took place in 2017, the achieved best object detection accuracy was 73.14%. All these measures range from 0 to 1 and the higher the better. For the PASCAL datasets, 8498 images are used to train a deep learning method called fully convolutional network (FCN) and only 67.5 % average precision was achieved [10]. From the visual results in [10], it is seen that even the input is the ground truth, the precision of the output is still not high. Since deep learning is a discrete mapping function and is not capable of generating continuous (infinite) outputs, the performance of deep learning in such cases will be reduced significantly. Consequently, no good results are expected when deep learning is used for image segmentation unless the resolution of the image is extremely small.

For better comparison, we list all the competitions mentioned above in Table 1 with their critical data information and best accuracies achieved so far. As can be seen, the performance of deep learning is inversely proportional to the number of outputs with the assumption that sufficient inputs are available. Similar to Shannon theorem, a theorem about the proportion of the number of the inputs and the number of the outputs should be studied to guide the performance of the deep learning.

Table 1. Comparisons of the best accuracies achieved by deep learning in international competitions with the critical data information

| Competitions | Number of inputs | Number of outputs | Best accuracy |
|---|---|---|---|
| Voice recognition | 6300 | 8 | 83.5% [2] |
| Digits recognition | 60000 | 10 | 99.77% [4] |
| Traffic sign recognition | 39209 | 43 | 99.46% [5] |
| Chinese handwriting recognition | 3900000 | 7356 | 97.39% [6] |
| ImageNet image classification | 1281167 | 1000 | 93.34% [7] |
| ImageNet object detection | 1281167 | ≈infinity | 73.14% [8] |
| PASCAL VOC image segmentation | 8498 | ≈infinity | 67.5 % [9];79.5%[10] |

### 3. Deep learning for biomedical image segmentation

In theory, deep learning does not have the ability to achieve an accurate image segmentation result since it simulates human brains to classify the objects based on approximation information instead of the exact information of the object as analyzed in the above sections. However, there are many reported research work about the unprecedented segmentation accuracy achieved by deep learning. Let us look into the details of how these researchers made it. In [11], the author who had won three international competitions [5-7] claimed that his deep neural network was trained with 3 million pixels with only 2 outputs: membrane denoted as 1 and non-membrane denoted as 0. The output is a prediction map which looks the same as the probability map generated by another competitor [12] and the popular method to generate the probability map is described in [13]. As can be seen, the prediction map is generated based on the features of the image instead of the pixels of the image. Therefore, deep learning works on the feature level instead of the pixel level for image segmentation. Accordingly, deep learning could not achieve pixel level accuracy without further processing. It is seen from the illustration in [14], there were serious segmentation errors in the final segmentation result by the deep learning method proposed in [11]. In another object detection competition [15], the same author of [11] used his deep learning method to detect mitosis in breast cancer histology images. At this time, the achieved F-measure accuracy was only 78.2%. So far, the author did not disclose any related source codes for other researchers to reproduce the segmentation accuracy or to verify the proposed deep learning method, which indicates that the international competitions based on the submitted results might not be reliable [16].

In [17], the authors gave a comprehensive survey of deep learning in cardiology and concluded that a complete theoretical understanding of deep learning is not yet available. Therefore, the authors claimed that successful application of artificial intelligence (AI) in the medical field relies on achieving interpretable models and big datasets. Unfortunately, deep learning is a black box instead of an interpretable model. Because of the black box property of deep learning, its performance might vary significantly in segmenting the same images with different implementations. We take the segmentation of the left ventricle (LV) by deep learning methods as a typical example and summarize the accuracies achieved by deep learning methods [18-47] based on public magnetic resonance image (MRI) and computational tomography (CT) datasets in Table 2. For the most popular dataset (2009 MICCAI challenge), the best accuracy achieved so far by deep learning based methods was 0.94 [21], where the deep learning was used to yield an approximation segmentation and then deformable model was used to refine the final segmentation result. The results in [21] verified the point that deep learning does not have the ability to achieve an accurate image segmentation result. However, the reported accuracy in [21] seems to be exaggerated. The inferred shape by deep learning is used as initialization contour for the evolvement of the deformable model. However, the accuracy of the deformable model is determined by the generated edge map instead of its initialization contour. Therefore, the final accuracy of the combined method should be similar to that of the deformable model based methods. On the contrary, the reported accuracy was much better than those reported by the deformable model based methods. In and after 2018, the reported accuracy on the 2009 MICCAI challenge datasets have dropped to 0.91 [27] and 0.9 [28].

Table 2. Comparison of the accuracies achieved by deep leaning methods on public left ventricle datasets (**Please note that the results in Table 2 should not be trusted fully since the accuracy is computed by the authors themselves or by the organizers based on the submitted results without verifying the corresponding source codes.**)

| Datasets\Performance | DICE similarity coefficients achieved in or before 2017 | DICE similarity coefficients achieved in or after 2018 |
|---|---|---|
| Datasets from York University(MRI dataset) | 0.75 [18]; 0.91[19] | NA |
| Sunnybrook 2009 MICCAI challenge (MRI dataset) | 0.93 [19];0.91[20];0.94[21];0.88[22];0.92 [23];0.85 [24];0.88[25];0.9 [26] | 0.91[27];0.9[28] |
| ACDC challenge STACOM 2017 (MRI dataset) | 0.86[29];0.97[30];0.95[31];0.96[32];0.95[33];0.95[34];0.95[35];0.92[36];0.82[37];0.94[38] | 0.92 [39-40] |
| MMWHS Challenge STACOM 2017 (CT dataset) | 80.9[41];90.8[42] | 79.3 [43] |
| STACOM 2011 challenge(MRI dataset) | 0.92 [44]; NA [45] | NA |
| STACOM 2013 challenge(MRI dataset) | 0.82 [46] | NA |
| STACOM 2018 challenge(MRI dataset) | NA | 0.85 [47] |

In 2017, statistical atlases and computational modeling (STACOM) of the heart workshop held two challenges: automated cardiac diagnosis challenge (ACDC) and multi-modality whole heart segmentation (MMWHS) challenge. The same as other international challenges or competitions, the competition teams were requested to submit their segmentation results to evaluate the accuracy of their methods. For the first time, deep learning based methods achieved the unprecedented accuracy by many teams, e.g. 0.97 [30], 0.96 [32] and 0.95 [32, 33-35]. One thing in common about the deep learning methods designed by these teams is that deep learning is the sole technique for segmentation and no other segmentation techniques were used for refinement. The organizer thus claimed that the problem of LV segmentation has been solved. Many of the teams had announced that they disclosed their codes or implementation details via the Gitbub. However, the author's team did not find any useful thing from their disclosed codes and they did not respond to the email either. Many other researchers could not reproduce similar accuracies either. For instance, the achieved accuracy on the ACDC datasets dropped from 0.95 to 0.92 in 2018 [39]. The achieved accuracy on the MMWHS datasets dropped from 0.91[42] to 0.79 [43] in 2018. In the latest STACOM 2018 challenge, the achieved accuracy for LV segmentation dropped to 0.85 [47]. As can be seen, the international competitions based on the submitted results are not reliable [16]. The reasons that made the accuracies achieved during STACOM 2017 so high include that (1), it is easy to improve the accuracy of the submitted results semi-automatically or manually; (2), The winners of the international challenges have a bright career future, which allure them to be dishonest. (3), most of them believe in that deep learning could actually reach that high accuracy even they did not achieve it during the practical implementations. Their blind faith in deep learning mainly comes from the great success of deep learning in classifying limited categories. As pointed out in this paper, deep learning could only give an approximation or a rough predication map instead of an accurate segmentation result for image segmentation.

**4. Comparison of deep learning methods with traditional methods**

Firstly, we compare the deep learning based methods with the slope difference distribution (SDD) threshold selection method proposed in [48]. The performance of SDD is affected by its parameters, the bandwidth $B$ and the fitting number $N$ significantly [48]. The segmentation results of selecting the bandwidth as 5 and the fitting number as 15 were published in [49] while its source codes had been made publicly accessible in 2017. Without notifying the reviewers that the source codes were available, the submitted paper had been rejected by many journals with almost the same reason that deep learning was better. However, the achieved accuracy by SDD in [49] is 0.9246 which is higher than most reported accuracies by deep learning methods. When the parameters of SDD are changed to 11 and 11 respectively, the segmentation accuracy could be improved to 0.95, which is greater than the accuracies reported by all deep learning methods on the same public datasets [50]. However, the selection rules of SDD become much more complex with a higher bandwidth and a lower fitting number. It seems that many reviewers could not accept the fact that deep learning is not the best image segmentation technique. It is not known why they believe in deep learning so deeply? Maybe they are overwhelmed by the great success of deep learning in other fields. So far, we only found two deep learning methods [51-52] that have disclosed their source codes and their trained models [53-54]. However, the reproduced accuracy is significantly lower than their claimed accuracy in their papers as shown in Table 3. On the contrary, SDD based method [55] achieved significantly better accuracy with freely available MATLAB source codes [56] for verification.

Table 3. Comparison of DICE similarity coefficients achieved by the disclosed codes of two deep learning methods and the SDD method on the ACDC dataset.

| Validation\Methods | Deep learning [51,53] | Deep learning [52,54] | SDD [55,56] |
|---|---|---|---|
| Published accuracy | 94% [51] | 94% [52] | **95.44% [55]** |
| Source codes accuracy | 90.28% [53] | 87.13% [54] | **95.44% [56]** |

Secondly, we compare the deep learning based methods with other segmentation methods in segmenting different medical images. The quantitative comparisons are shown in Table 4. The achieved Jaccard Index accuracy by the deep learning method for ventricle segmentation in [47] was only 0.77 which is lower than that of a tracking method [57]. Deep learning was combined with surface evolution to segment CT livers in [58]. Deep learning was combined with graph cut to segment the CT livers in [59]. Even combined with deep learning for prediction, the reported liver segmentation accuracy was still lower than the previously reported liver segmentation accuracy by the non-deep-learning methods [60-61]. For the liver tumor detection, the reported accuracy by deep learning was only 69% [62] compared to 88.9% computed by a non-deep-learning method [63]. Deep learning was also used for cell segmentation [64] and cell tracking [66]. Deep learning was combined with watershed for cell segmentation in [58] and the reported accuracy was only 86%. In addition, it was reported in [65] that the accuracy of cell segmentation by support vector machine was more robust than deep learning. In [66], deep learning was used for cell tracking and it was reported that only one cell could be tracked, which is much more inefficient than traditional cell tracking methods [67-68].

The cases described above are only the tip of the iceberg, there are also many cases in which deep learning performed very poorly or it was outperformed by other methods. However, it is usually difficult to publish these results since the traditional methods may be lack of originality for publishing. In many cases, the claimed accuracy in the published paper and the reported accuracy in international competitions might make people think that it is the inability of the researcher instead of inability of deep learning for the failure of a segmentation task. Almost no authors that use deep learning for image segmentation would like to disclose enough data, source codes and the required information to reproduce their results. Besides the deliberate exaggeration, there might be mistakes that caused the reported accuracy to be unreliable.

Table 4. Comparisons of the accuracies of deep leaning methods and non-deep-learning methods in segmenting the medical images (**Please note that the results in Table 4 should not be trusted fully since the accuracy is computed by the authors themselves without verifying the corresponding source codes.**)

| Competitions | Deep learning methods | non-deep-learning methods |
|---|---|---|
| MRI Ventricle segmentation | Jaccard Index 0.77 [47] | Jaccard Index 0.84 [57] |
| CT liver segmentation on MICCAI-Sliver07 test set | Score 79.3% [58]; 77.8% [59] | Score 79.6% [60] |
| CT liver segmentation on 3Dircabd database | VOE 9.36±3.34 [59] | VOE 9.15±1.44 [61] |
| Liver tumor segmentation | DICE 69% [62] | DICE 88.9% [63] |
| Cell segmentation | OR 0.83% [64] | OR 0.69% [65] |
| Cell tracking | Capable of tracking one cell at a time [66] | Capable of tracking hundreds of cells simultaneously [67-68] |
| Please note: Jaccard Index, DICE and Score: the higher the better; VOE and OR: the lower the better. | | |

## 5. Discussion

The fetishistic devotion to deep learning is mainly caused by its overhype. This overhype is partially caused by the international challenges. Deep learning monopolized the champions of all the international challenges. Ironically, the organizers of the international challenges only evaluate the submitted results instead of the source codes or the software to rank the competitors. As a result, deep learning won almost all the international image segmentation challenges. According to Gary Marcus, the professor at New York University, "Deep learning systems are most often used as classification system in the sense that the mission of a typical network is to decide which of a set of categories (defined by the output variables on the neural network) a given input belongs to." [69]. According to our analysis in this paper, deep learning is a discrete, but not continuous regression model. As a discrete non-linear mapping function, the performance of deep learning will be affected significantly when the number of the outputs approaches infinity. Thus, deep learning could not achieve pixel-level accuracy for image segmentation as traditional methods can.

Compared to the regression model, deep learning has the major advantage that deep learning is more robust in estimating the discrete response variables for the linear indivisible data. Up to now, deep learning has been recognized as the most powerful classification tool. However, deep learning also has the following limitations.

(1), deep learning determines a set of parameters instead of the exact mathematical functions to map the input variables into the output variables. Without the exact mathematical functions, not only deep learning could not be explained reasonably, but also the output variables could not be determined as continuous response variables. Consequently, deep learning could only map sufficient inputs into a finite number of categories. When the number of output is infinite or close to infinity, the performance of deep learning will be affected significantly.

(2), Lack of the exact mathematical mapping, deep learning could not yield the ideal outputs even when it is tested with the inputs used during the training process or tested with the ground truth.

(3), Lack of the exact mathematical mapping, deep learning is only powerful for the classification applications, where sufficient input variables are available while the number of output variables is comparatively limited. In information theory, there is Shannon theorem that tells the maximum rate at which information can be transmitted over the channel with the specified bandwidth in the presence of noise. In deep learning, there is no theorem to determine the maximum number of the output variables when the input variables are known, which indicates that deep learning is immature.

(4), the training process of deep learning relies on large amounts of human-annotated data to determine the parameters. However, when different annotated training data are used, the performance of deep learning may vary significantly.

(5), there is no known efficient training algorithms with provable guarantees for deep learning and its success relies on descent algorithms, such as coordinate, gradient or stochastic gradient descent [2].

(6), since there are no guaranteed rules for the training process of deep learning, deep learning is easy to be overhyped or utilized by people who conduct false advertising for their deep learning based techniques or products.

## 6. Conclusion

In conclusion, deep learning could only yield a semantic-level predication map for image segmentation and relies on other image segmentation techniques to obtain the pixel-level accuracy. When the number of the outputs approaches infinity, the approximating ability or the classification ability of deep learning will go down significantly. Deep learning became so hot not because it is perfect or omnipotent, but because no alternative techniques are available yet. Excessive overhype of deep learning is partially caused by some international challenges that have significant scientific impact, but do not evaluate the competitors' algorithms strictly.